\documentclass[10pt,conference]{IEEEtran}

\newif\ifanonymize
\ifanonymize \def\hl#1{}\else
\let\hl\relax
\fi

\IEEEoverridecommandlockouts
\usepackage{ascmac,amsmath}
\usepackage[dvipdfmx]{graphicx}
\usepackage[utf8]{inputenc}
\usepackage{latexsym}
\usepackage{multirow}
\usepackage{url}
\usepackage{xcolor} 
\usepackage{amsfonts}

\usepackage{cite}
\usepackage{ascmac}
\usepackage{subfigmat}

\begin{document}
\title{\textsc{NeuRecover}: Regression-Controlled Repair of Deep Neural Networks with Training History}
\hl{
\author{
\IEEEauthorblockN{Shogo Tokui}
\IEEEauthorblockA{\textit{Fujitsu Limited}\\
Kawasaki, Japan \\
tokui.shogo@fujitsu.com}
\and
\IEEEauthorblockN{Susumu Tokumoto}
\IEEEauthorblockA{\textit{Fujitsu Limited}\\
Kawasaki, Japan \\
tokumoto.susumu@fujitsu.com}
\and
\IEEEauthorblockN{Akihito Yoshii}
\IEEEauthorblockA{\textit{Fujitsu Limited}\\
Kawasaki, Japan \\
yoshii.akihito@fujitsu.com}
\and
\IEEEauthorblockN{Fuyuki Ishikawa}
\IEEEauthorblockA{\textit{National Institute of Informatics}\\
Tokyo, Japan \\
f-ishikawa@nii.ac.jp}
\and
\IEEEauthorblockN{Takao Nakagawa}
\IEEEauthorblockA{\textit{Fujitsu Limited}\\
Kawasaki, Japan \\
nakagawa-takao@fujitsu.com}
\and
\IEEEauthorblockN{Kazuki Munakata}
\IEEEauthorblockA{\textit{Fujitsu Limited}\\
Kawasaki, Japan \\
munakata.kazuki@fujitsu.com}
\and
\IEEEauthorblockN{Shinji Kikuchi}
\IEEEauthorblockA{\textit{Fujitsu Limited}\\
Kawasaki, Japan \\
skikuchi@fujitsu.com}
}}

\maketitle

\thispagestyle{plain}
\pagestyle{plain}

\begin{abstract}
    Systematic techniques to improve quality of deep neural networks (DNNs) are critical given the increasing demand for practical applications including safety-critical ones. The key challenge comes from the little controllability in updating DNNs. Retraining to fix some behavior often has a destructive impact on other behavior, causing regressions, i.e., the updated DNN fails with inputs correctly handled by the original one. This problem is crucial when engineers are required to investigate failures in intensive assurance activities for safety or trust.
    
    Search-based repair techniques for DNNs have potentials to tackle this challenge by enabling localized updates only on ``responsible parameters'' inside the DNN. However, the potentials have not been explored to realize sufficient controllability to suppress regressions in DNN repair tasks.

    In this paper, we propose a novel DNN repair method that makes use of the training history for judging which DNN parameters should be changed or not to suppress regressions. We implemented the method into a tool called \textsc{NeuRecover} and evaluated it with three datasets. Our method outperformed the existing method by achieving often less than a quarter, even a tenth in some cases, number of regressions. Our method is especially effective when the repair requirements are tight to fix specific failure types. In such cases, our method showed stably low rates ($<$2\%) of regressions, which were in many cases a tenth of regressions caused by retraining.
\end{abstract}


\section{Introduction}
\label{sec:introduction}
Deep neural networks (DNNs) have recently been used in systems for applications such as speech recognition \cite{hinton2012deep}, machine translation \cite{cho2014learning}, object detection \cite{ren2015faster}, sentiment analysis \cite{tang2015document}, and face recognition \cite{schroff2015facenet}. They have also been used in safety-critical industrial applications such as medical diagnosis \cite{litjens2017survey}, autonomous driving \cite{chen2015deepdriving}, and aircraft collision avoidance systems \cite{julian2016policy}.
However, engineering methodologies for the development, quality assurance, and operation of DNN-based systems were first discussed only a few years ago \cite{amershi2019software,ishikawa2019engineers,hamada2020guidelines}, and there are serious concerns about quality and continuous maintenance and improvement, in particular.

DNNs and other machine-learning-based software are referred to as Software 2.0 \cite{karpathy2017software}, which means software consists of enormous number of interconnected parameters and the behavior is derived in a data-driven way via training. This characteristic introduces a challenge in continuous improvement. Specifically, updates by retraining affect the whole behavior of the DNN. In other words, we do not have control to localize the changes to have limited impact on the specific behavior. This nature is even said as ``Changing Anything Changes Everything''  \cite{sculley2015hidden}.

As DNNs have been applied to more safety-critical or quality-sensitive domains, suppressing regressions is increasingly crucial. For example, stakeholders are curious about whether there are unacceptable mistakes with high risks that can lead to serious hazards or distrust by stakeholders. In such a case, engineers are required to have costly activities to check failed cases and give some explanation. Regressions lead to high cost of redoing such activities. This is true even if the total accuracy is improved as the combined effect of improvements and regressions.

Traditional software engineering techniques have great potentials to tackle problems in DNNs. 
Techniques for automated program repair have potentials to realize effective methods for automated DNN repair. Many techniques have been proposed to fix programs, especially the ``Generate and Validate'' technique, which has evolved significantly in the last decade and has been highly successful in fixing simple faults \cite{le2012systematic,saha2017elixir,noda2020experience}.

There has already been a study to apply the search-based approach for the DNN repair problem. Sohn et al. proposed Arachne \cite{arachne2019Sohn}, a method for turning misclassified data into correctly classified data by changing the parameters (weights) of the DNN model in an exploratory manner.
This method consists of fault localization to identify the weights causing misclassification and particle swarm optimization \cite{pso1995James,pso2007Andreas} to find weight values that will reduce the error. 
However, because the fault localization of Arachne identifies the target weights by considering only their impact on the misclassified data, there is a high possibility that the method will turn correctly classified data into misclassified data. Arachne thus shares the common problem of regressions as retraining.

In this paper, we propose a novel DNN repair technique, \textsc{NeuRecover}, that suppresses regressions by using the training history in the fault localization step.
The basic idea of \textsc{NeuRecover} is to find the point in the training history when the model correctly classified a certain data sample that is now misclassified, and then to identify weights that can safely correct the misclassification, by comparing the past model with the current model.
Specifically, \textsc{NeuRecover} identifies weights with the following properties: their values have changed significantly in the training process, and they do not affect the output for improved data (i.e., data that was first misclassified but then classified correctly in the training process), but they do affect the output for regressed data (i.e., data that was once correctly classified but then misclassified in the training process).
Then, by applying particle swarm optimization on the identified weights, \textsc{NeuRecover} can update the DNN model to obtain more improved data and less regressed data.

We experimentally evaluated \textsc{NeuRecover} with models with three DNN architectures by using three image classification datasets, GTSRB, CIFAR-10, and Fasihon-MNIST. \textsc{NeuRecover} outperformed the baseline method Arachne by achieving often less than a quarter, even a tenth in some cases, number of regressions. \textsc{NeuRecover} is especially suitable when the repair requirements are tight to fix specific failure types while avoiding regressions. In such cases, \textsc{NeuRecover} showed stably low rates ($<$2\%) of regressions, in many cases a tenth, at most a quarter, compared with retraining that tend to have large shuffling of success and failure cases.

The contributions of this paper are summarized as follows:
\begin{itemize}
    \item A novel DNN repair technique that suppresses regression by using the training history.
    \item An implementation of the technique, called \textsc{NeuRecover}, including algorithm improvements from Arachne, an existing method for search-based DNN repair.
    \item Experiments to investigate repair performance in the design space for search-based DNN repair methods.
    \item Experiments to investigate repair performance with loose and tight requirements, respectively, for search-based DNN repair methods as well as common retraining.
\end{itemize}

The remainder of this paper is structured as follows.
Section \ref{sec:background} describes DNNs and the existing technique, Arachne, as the background of this study.
Section \ref{sec:our_method} describes the proposed technique, \textsc{NeuRecover}.
Section \ref{sec:experiment} describes the evaluation experiment and discusses the proposed technique and its validity. 
Finally, section \ref{sec:relatedworks} describes related works, and section \ref{sec:summary} summarizes this paper and our future works.

\section{Background}
\label{sec:background}
This section describes DNNs as well as Arachne, an existing DNN repair technique that directly corrects the parameters (weights) of DNN models without retraining.

\subsection{Deep Neural Network (DNN)}
\label{subsec:dnn}
A DNN is a neural network composed of an input layer, an output layer, and two or more hidden layers.
In particular, a feedforward neural network (FFNN) is known as a primary neural network to solve classification problems.
An FFNN, propagates information through an input layer, a hidden layer, and an output layer, in order, and it outputs a prediction label for the input.

This section explains how to train a DNN model.
The model has the DNN architecture and parameter values for the hidden layer.
Each parameter value of the hidden layer is adjusted using training data.
For data $x$ given by the input layer, the hidden layer converts it to $o=wx+b$ via two parameters, a weight $w$ and a bias $b$; then, $x'=A(o)$ is outputted via a nonlinear derivative function $A$ called an activation function.
The output layer obtains the index of the largest element of the hidden layer's output and gives its prediction result.
The function representing the error between the expected and predicted labels for the data is called the loss function $L$.
For instance, the squared error is one such loss function.
A smaller loss indicates a better model.
In the training of a DNN model, the parameters are adjusted by using the error backpropagation method to reduce the loss \cite{rumelhart1986learning}.
The error backpropagation method executes the steepest descent method, in which it adjusts the weights to $w=w-\alpha \frac{\partial L}{\partial w}$ by using the learning rate $\alpha$.
When a model is trained for $n$ epochs, the error backpropagation method is repeated $n$ times.

The field of image recognition uses convolutional neural networks (CNNs).
A CNN is a DNN in which a convolutional layer for image processing is added to the hidden layer.
Whereas the convolutional layer propagates to the subsequent stage via one feature, which convolves part of the region of the neurons in the previous stage, the layer to which all the neurons in the previous and subsequent stages are connected is called the fully connected layer.
A basic CNN thus consists of four layers: an input layer, a convolutional layer, a fully connected layer, and an output layer.

A system using a DNN can build a model by training with data.
When erroneous behavior is detected during operation or testing of the system, the DNN model is modified by adding data and retraining.
However, retraining requires additional data to correct misclassified data, and it is not always possible to correct a model's output with added data.
Therefore, to correct a DNN model without retraining, techniques have been studied to correct misclassification by directly manipulating the values of the model's weight parameters.

\subsection{Search-Based DNN Repair}
\label{subsec:arachne}
Arachne is a proposed DNN repair technique that locally modifies a DNN without retraining by changing the model's parameters (i.e., the weights) in an exploratory manner \cite{arachne2019Sohn}.
It works by identifying a weight that induces misclassification in the trained model, adjusting the value of the weight by using particle swarm optimization \cite{pso1995James,pso2007Andreas}, and correcting the misclassification to the expected classification.

Arachne's repair process consists of two steps: fault localization and particle swarm optimization.
In fault localization, the gradient of each weight loss function in the DNN model and the output value of each layer are calculated as the impacts of the weights on misclassified data, and the weight causing misclassification is identified.
In particle swarm optimization, by applying a fitness function, the weight value specified by fault localization is optimized to increase the amount of misclassified data that can be correctly classified.
The details of fault localization and particle swarm optimization are described below.

\subsubsection{Fault Localization}
\label{subsubsec:arachne_localize}
Because a DNN model includes more than tens of thousands of weights, it is very expensive to adjust all weights at the same time by particle swarm optimization.
Accordingly, to narrow down the weights to be optimized, in the fault localization step Arachne focuses on weights connected to the final layer and then tries to identify those that have a large impact on misclassification.
It uses two methods to evaluate each weight's impact on a specific misclassification: (1) the {\bf gradient of the loss function}, which is used to adjust the weight and bias, and (2) the {\bf output value of forward propagation}, which indicates the activation of neurons during model training.

First, Arachne inputs a misclassified sample to the DNN model to be repaired.
Next, it obtains the values of the weight parameters in the final layer.
Finally, it ranks the weights by considering both the gradient of the loss function and the output value of forward propagation.
The gradient of the loss function is the value $\frac{\partial L}{\partial w}$ obtained by differentiating the loss function $L$ by the weight $w$; it is calculated as $\frac{\partial L}{\partial w} = \frac{\partial L}{\partial o} \frac{\partial o}{\partial w}$ by using the output $o$ of forward propagation in the final layer.
If the output of the $j$-th neuron in the last layer is $o_j$, then the gradient of the loss function with respect to the weight $w_{i,j}$ and the $i$-th neuron in the previous layer is calculated as $\frac{\partial L}{\partial w_{i,j}} = \frac{\partial L}{\partial o_j} \frac{\partial o_j}{\partial w_{i,j}}$.
The output value of forward propagation for weight $w_{i,j}$ is calculated by multiplying the output $o_i$ of the layer before the activation function's nonlinear conversion by the weight $w_{i,j}$, i.e., as $o_i \cdot w_{i,j}$.

The number of candidate weights selected according to the gradient of the loss function is determined to be $N_g$ in advance.
That is, the weights are sorted by their gradients, and the top $N_g$ weights are treated as candidates.
Finally, to extract the set of weights to be optimized in the next step, the Pareto front is calculated after performing multi-objective optimization with both the gradient of the loss function and the output values of forward propagation as objective functions.

\subsubsection{Patch Generation}
\label{subsubsec:arachne_optimize}
Arachne corrects a DNN model's misclassification by using particle swarm optimization for the weights specified by fault localization \cite{pso1995James,pso2007Andreas}.
Particle swarm optimization is known to be effective for optimization in a continuous space and is suitable for unrestricted weight modification in the range of real numbers.

Arachne expresses the particle positions in particle swarm optimization, with the set of weights specified by fault localization, as a vector $\vec{x}$.
The current particle vector $\vec{x}_t$ is updated using the velocity vector $\vec{v}_{t+1}$ via Equation \ref{eq:x} below.
The current velocity vector $\vec{v}_t$ is updated via Equation \ref{eq:v} by using the current particle vector $\vec{x}_t$; a particle vector $\vec{p}_l$, which takes the best fit value among the particle's previously observed values; a particle vector $\vec{p}_g$, which takes the best fit value for the whole group; and a uniform random number $U(\phi)\,(0 \le U(\phi) \le \phi)$.
Here, $\phi_1$ and $\phi_2$ control the convergence of particles in a group without setting explicit velocity boundaries for local and global components, respectively.
The value $\chi$, which is called a constriction factor, is calculated from $\phi_1$ and $\phi_2$ via Equation \ref{eq:chi}.
Arachne uses the same values for $\phi_1$ and $\phi_2$.
It extracts the particle vector's initial value $\vec{x}_0$ from a normal distribution determined by the distribution of weights, with the initial velocity $\vec{v}_0$ set to $\vec{0}$.

\begin{align}
    \label{eq:x}
    \vec{x}_{t+1} & \leftarrow \vec{x}_t + \vec{v}_{t+1} \\
    \label{eq:v}
    \vec{v}_{t+1} & \leftarrow \chi (\vec{v}_t + U(\phi_1)(\vec{p}_l - \vec{x}_t) + U(\phi_2)(\vec{p}_g - \vec{x}_t)) \\
    \label{eq:chi}
    \chi          & \leftarrow \frac{2}{\phi - 2 + \sqrt{\phi^2 - 4 \phi}}, \, {\rm where} \, \phi = \phi_1 = \phi_2
\end{align}

According to the fitness function given in Equation \ref{eq:arachne_fitness} below, $\vec{p}_l$ and $\vec{p}_g$ in Equation \ref{eq:v} use the particles with the best fitness values among the particles observed in the past.
Here, $I_{\textrm{neg}}$ is a set of misclassified samples, and $I_{\textrm{pos}}$ is a set of randomly selected samples that were correctly classified.
Lastly, $N_{\textrm{patched}}$ is the number of data instances in $I_{\textrm{neg}}$ that were changed from misclassification to the expected classification, whereas $N_{\textrm{intact}}$ is the number of data instances in $I_{\textrm{pos}}$ that were not changed from the expected classification to misclassification.

\begin{eqnarray}
    \label{eq:arachne_fitness}
    \textit{fitness} = \frac{N_{\textrm{patched}} + 1}{L(I_{\textrm{neg}}) + 1} + \frac{N_{\textrm{intact}} + 1}{L(I_{\textrm{pos}}) + 1}
\end{eqnarray}

\section{\textsc{NeuRecover}: DNN Repair with Training History}
\label{sec:our_method}
Arachne identifies the parameters (weights) that affect misclassification and searches for weight values that reduce the error by using particle swarm optimization.
However, Arachne has a potential risk of introducing new misclassification into the model while correcting some misclassification.
We consider the cause of this problem to be that Arachne uses only misclassified data as information for fault localization and does not consider correctly classified data.
In other words, if some of the weights identified by the fault localization step affect data that was correctly classified, then the patch generation step may turn correctly classified samples into misclassified samples.
To solve this problem, we made the following two assumptions:
(1) It should be relatively easy to correct misclassification if a misclassified sample was once classified correctly during the training process.
(2) If we identify weights that impact the results for correctly classified samples, then we can reduce data regression by avoiding changes to the values of those weights.

\begin{figure}
    \centering
    \includegraphics[width=0.9\linewidth]{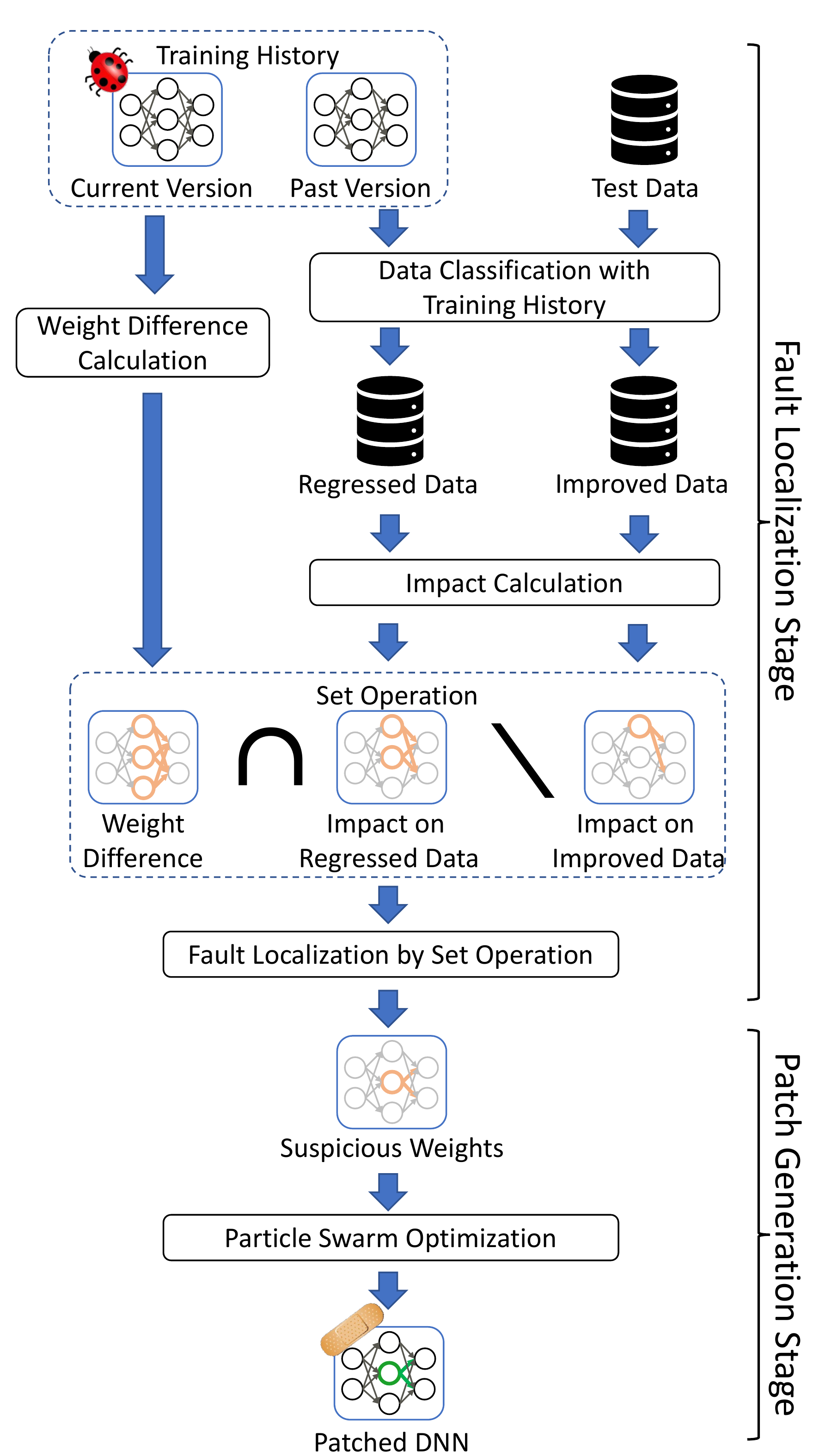}
    \caption{Overview of \textsc{NeuRecover}}
    \label{fig:overview}
\end{figure}

Hence, we propose \textsc{NeuRecover}, which is a novel DNN repair technique that uses fault localization with the training history.
In this technique, the fault localization detects improved data and regressed data in the training process.
Here, we use the term \textsl{improved data} to refer to data that changed from misclassified to correctly classified and the term \textsl{regressed data} to refer to data that changed from correctly classified to misclassified.
Then, as illustrated in Figure \ref{fig:overview}, \textsc{NeuRecover} identifies weights that have changed significantly during the training process and do not affect improved data but only affect regressed data (Fault Localization Stage), and optimizes the localized weights by particle swarm optimization as in Arachne (Patch Generation Stage).
The fault localization step using the training history is executed in the following three steps.

\begin{description}
    \setlength{\itemindent}{9pt}
    \setlength{\labelsep}{15pt}
    \item[Step\,i] Data classification with training history
    \item[Step\,i\hspace{-1pt}i] Impact calculation
    \item[Step\,i\hspace{-1pt}i\hspace{-1pt}i] Fault localization by set operation
\end{description}

In Step\,i, \textsc{NeuRecover} uses the training history to detect regressed and improved data in the test data.
In Step\,i\hspace{-1pt}\nolinebreak[0]i, it calculates the impact on the regressed data, the impact on the improved data, and the difference between the weights.
In Step\,i\hspace{-1pt}\nolinebreak[0]i\hspace{-1pt}\nolinebreak[0]i, it identifies the sets of weights affecting the regressed data, the improved data, and the difference between the weights, and it uses a set operation to localize these weights.
Finally, in patch generation stage, \textsc{NeuRecover} corrects the localized weights of the DNN model by using particle swarm optimization.
We describe the details of each step below.

\subsection{Step\,i: Data Classification with Training History}
\label{subsec:deg_imp_from_history}

First, \textsc{NeuRecover} detects regressed data and improved data from the training history.
To repair of a model that was trained for $n$ epochs, it uses the weights of models $M_n$ and $M_{n-k}$ in the training history, where $M_i$ denotes the model that was trained for $i (1 \le i \le n)$ epochs.

For each model, \textsc{NeuRecover} examines the predictions of the dataset.
It classifies data that changed from the expected classification by $M_{n-k}$ to misclassification by $M_n$ as regressed data, and data that changed from misclassification by $M_{n-k}$ to the expected classification by $M_n$ as improved data.
In our evaluation experiment described in this paper, the datasets were classified using the models $M_{n-1}$ and $M_n$ ($k=1$).

\subsection{Step\,i\hspace{-1pt}i: Impact Calculation}
\label{subsec:calc_impact}
Next, \textsc{NeuRecover} calculates five impacts to identify the weights to be corrected: the weight difference, $w_{\textrm{diff}}$; the backward impact on regressed data, $\textit{back}_{\textrm{reg}}$; the forward impact on regressed data, $\textit{fwd}_{\textrm{reg}}$; the backward impact on improved data, $\textit{back}_{\textrm{imp}}$; and the forward impact on improved data, $\textit{fwd}_{\textrm{imp}}$.
In this study, the backward impact is given by the gradient of the loss function, and the forward impact is given by the output value of forward propagation.

We assume that a large weight difference $w_{\textrm{diff}}$ is the cause of changes in the prediction results.
\textsc{NeuRecover} thus obtain a weight array $w_n$ from $M_n$ and a weight array $w_{n-k}$ from $M_{n-k}$ and calculates $w_{\textrm{diff}}=w_n-w_{n-k}$.

The backward impact $\textit{back}$ is calculated as $\textit{back} = \frac{\partial L}{\partial w_{i,j}} = \frac{\partial L}{\partial o_j} \frac{\partial o_j}{\partial w_{i,j}}$ for the localized weights $w_{i,j}$, which connect the $j$-th neuron of the previous layer and the $i$-th neuron of the target layer, and the output activation value $o_j$ of the $j$-th neuron of the target layer.
$\textit{back}$ thus depends on the loss function $L$, the neuron output $o$, the weights $w$, and the inputs.
\textsc{NeuRecover} obtains $\textit{back}_{\textrm{reg}}$ as the backward impact on regressed data and $\textit{back}_{\textrm{imp}}$ as the backward impact on improved data.

The forward impact $\textit{fwd}$ is calculated as $\textit{fwd} = o_i \cdot w_{i,j}$ from the output activation value $o_i$ of the $i$-th neuron of the previous layer and the weights $w_{i,j}$.
It thus depends on the weights $w$ and the inputs.
\textsc{NeuRecover} obtains $\textit{fwd}_{\textrm{reg}}$ as the forward impact on regressed data and $\textit{fwd}_{\textrm{imp}}$ as the forward impact on improved data.

Note that \textsc{NeuRecover} localizes the weights of all fully connected layers, not just the last layer of the DNN model.
That is, it computes the backward and forward impact for each layer of the DNN model.
In this paper, for hypothesis verification in the initial stage, only the fully connected layers are examined, but in the future, we will consider correcting the weights of the convolutional layers, as well.

\subsection{Step\,i\hspace{-1pt}i\hspace{-1pt}i: Fault Localization by Set Operation}
\label{subsec:our_localize}

In the last step, the weights are sorted for each of the five impacts $w_{\textrm{diff}}$, $\textit{back}_{\textrm{reg}}$, $\textit{fwd}_{\textrm{reg}}$, $\textit{back}_{\textrm{imp}}$, and $\textit{back}_{\textrm{imp}}$.
Then, the five corresponding sets $W_{\textrm{diff}}$, $B_{\textrm{reg}}$, $F_{\textrm{reg}}$, $B_{\textrm{imp}}$, and $F_{\textrm{imp}}$ of the top $N_g$ weights are obtained.
Finally, the weights specified by the set operation in Equation \ref{eq:set} are defined as the target of DNN model repair.

\begin{eqnarray}
    \label{eq:set}
    W_\textrm{localized} = (B_{\textrm{reg}} \cap F_{\textrm{reg}}) \cap W_{\textrm{diff}} \backslash (B_{\textrm{imp}} \cap F_{\textrm{imp}})
\end{eqnarray}

By Equation \ref{eq:set}, \textsc{NeuRecover} identifies a localized set $W_\textrm{localized}$ of weights that have a large difference and affect the regressed data, while excluding weights that affect the improved data.
We consider weights that affect the regressed data and improved data to have large values for both the backward and forward impacts.
Therefore, the weights that affect the regressed data are given by $B_{\textrm{reg}} \cap F_{\textrm{reg}}$, and the weights that affect the improved data are given by $B_{\textrm{imp}} \cap F_{\textrm{imp}}$.
The set operation in Equation \ref{eq:set} thus suppresses data regression during fault localization and repair of the DNN model.

\subsection{Patch Generation Stage}
\label{subsec:our_optimize}
In the patch generation stage, \textsc{NeuRecover} corrects the localized weights by using particle swarm optimization as described in section \ref{subsubsec:arachne_optimize}.
For this paper, however, we changed the fitness function and the samples chosen from the data that was correctly classified.

The fitness function used in Arachne is positively proportional to the number of corrected data instances and intact data instances.
This means that the fitness value depends on the number of misclassified data instances and sampled correctly classified data instances, and is considered to be oversensitive to the absolute amount of these data instances.
To mitigate the sensitivity and obtain stable results, we changed the fitness function to use a relative amount of misclassified and correctly classified data before and after running the method.
$\alpha$ is a hyper-parameter to adjust the degree of regression suppression.

\begin{eqnarray}
    \label{eq:our_fitness}
    \textit{fitness} = \frac{N_{\textrm{patched}} / |I_{\textrm{neg}}| + 1}{L(I_{\textrm{neg}}) + 1} + \alpha \cdot \frac{N_{\textrm{intact}} / |I_{\textrm{pos}}| + 1}{L(I_{\textrm{pos}}) + 1}
\end{eqnarray}

Note also that Arachne randomly selects the samples of correctly classified data.
However, the larger the mean square error between the predicted and correct values for the correctly classified data is, the closer the that data is to the classification boundaries.
We consider prevention of the regression of correctly classified data that is close to the classification boundaries to also prevent regression of other correctly classified data.
Accordingly, \textsc{NeuRecover} selects the samples of correctly classified data in order of the error size.

\section{Evaluation}
\label{sec:experiment}
In this section, we describe an evaluation experiment of the proposed technique \textsc{NeuRecover}.

In the experiment, we compared three methods, \textsc{NeuRecover}, Arachne, and retraining, and evaluated the design validity of \textsc{NeuRecover} on three datasets and three model architectures for image classification.

\subsection{Experiment Setup}
\label{subsec:setup}
\subsubsection{Model Architectures and Datasets}
\label{subsubsec:archdata}
To avoid biasing the evaluation toward any particular model, we tried nine combinations of model architectures and datasets.
We prepared three different model architectures: 8-layer CNN (8CN), VGG16 (V16), and VGG19 (V19).
The 8-layer CNN consists of six convolutional layers and two fully connected layers, the VGG16 consists of 13 convolutional layers and three fully connected layers, and the VGG19 consists of 16 convolutional layers and two fully connected layer.
We also prepared three different image classification datasets: GTSRB (GT) \cite{gtsrb}, CIFAR-10 (C10) \cite{cifar10} and Fashion-MNIST (FM) \cite{fmnist}.

\subsubsection{Data Split and Categorization}
\label{subsubsec:categorization}
Each dataset was divided into three categories: \textit{train}, \textit{repair}, and \textit{test}. A \textit{repair} category was specially defined, being separated from a \textit{train} category for the debugging process.

The \textit{train} category is used at a training prior to the fault localization steps described in the section \ref{sec:our_method}. During the fault localization steps, the data classification proceeds with data samples taken from the \textit{repair} category. After the optimization process has been completed, a repaired model is evaluated using the \textit{test} category samples.

Since the datasets are originally split into two categories, train and test, we split the original train data into two new categories, \textit{train} and \textit{repair}, as in $K$-fold cross validation.
Multiple patterns of a separation between \textit{train} and \textit{repair} can be defined. Let $K \in\mathbb{N}$ as the number of patterns. The whole part of the original train data can be divided into $K$ segments. We define a \textit{repair} category as one of the $K$ segments and define a \textit{train} category as data samples basically including $K-1$ segments; therefore, $K$ possibilities of combinations of the \textit{repair} and the \textit{train} category can be considered.

Models trained only on the \textit{train} category data are regarded as faulty baseline models, and the models are subject to correction in each technique with the \textit{repair} category data.
The data classification results in the baseline models with the \textit{test} category data are shown in Table \ref{tab:data_category}. We have chosen the $K = 5$ condition.\footnote{We experimented with 4 patterns of the 5 combinations for $K=5$ segments due to a defect in the experiment source code and time constraints.}
The \#pos, \#neg, \#reg, and \#imp are the mean number of correctly classified, misclassified, regressed, and improved data for $K = 5$ patterns, respectively.
The \#pos and the \#neg are calculated from the classification results by a model trained until the last epoch (i.e. the 10th epoch).
On the other hand, the \#reg and the \#imp show the change of the classification results at the last epoch in comparison of the one before epoch (i.e. the 9th epoch). 

\begin{table*}[t]
    \centering
    \caption{The classification results of baseline models}
    \label{tab:data_category}
    \begin{tabular}{|l|l|r|r|r|r|r|r|}
        \hline
        Datasets & Model Arch. & Epochs & ACC & \#pos & \#neg & \#reg & \#imp \\
        \hline \hline
        GTSRB & 8-layer CNN & 10 & 96.247 & 12156.0 & 474.0 & 194.3 & 166.8 \\
        \hline
        GTSRB & VGG16 & 10 & 89.733 & 11333.3 & 1296.8 & 319.3 & 344.0 \\
        \hline
        GTSRB & VGG19 & 10 & 53.830 & 6798.8 & 5831.3 & 399.0 & 441.3 \\
        \hline
        
        CIFAR10 & 8-layer CNN & 10 & 74.580 & 7458.0 & 2542.0 & 734.3 & 906.8 \\
        \hline
        CIFAR10 & VGG16 & 10 & 80.880 & 8088.0 & 1912.0 & 487.5 & 477.0 \\
        \hline
        CIFAR10 & VGG19 & 10 & 57.738 & 5773.8 & 4226.3 & 326.0 & 350.5 \\
        \hline
        
        Fashion-MNIST & 8-layer CNN & 10 & 90.555 & 9055.5 & 944.5 & 256.5 & 251.3 \\
        \hline
        Fashion-MNIST & VGG16 & 10 & 91.110 & 9111.0 & 889.0 & 153.0 & 170.5 \\
        \hline
        Fashion-MNIST & VGG19 & 10 & 83.850 & 8385.0 & 1615.0 & 179.8 & 194.5 \\
        \hline
    \end{tabular}
\end{table*}

\subsubsection{Competitors}
\label{subsubsec:competitors}
The experiment compared our proposed method with Arachne and retraining.
We implemented the experimental code for Arachne according to the Arachne paper because its implementation was not published.
Retraining is a method that attempts to improve the model by adding data that is not included in the training dataset and training again with that data.
Developers and maintainers of ML systems generally use it for repairing their ML model when they find faults in the model.
In the experiment, we call ``retraining'' the same epochs training as the baseline model from the initial state with the \textit{train} category data, misclassified data in \textit{repair} category, and sampled correctly classified data in the \textit{repair} category.
The ``retraining'' allows us to observe only the effects of the increased \textit{repair} category data in training.
Note that since the baseline model has not been trained for a sufficient number of epochs, additional training of the models will increase the accuracy with or without the \textit{repair} category data.

\subsubsection{Metrics}
\label{subsubsec:metrics}
The DNN repair performance was evaluated in terms of the accuracy, repair rate, and break rate, which are given by the following equations.
\begin{eqnarray}
    \textit{Accuracy (ACC)} = |I_{\textrm{pos}}| / |I_{\textrm{all}}| \times 100 \nonumber \\
    \textit{Repair Rate (RR)} = |I_{\textrm{imp}}| / |I_{\textrm{neg}}| \times 100 \nonumber \\
    \textit{Break Rate (BR)} = |I_{\textrm{reg}}| / |I_{\textrm{pos}}| \times 100 \nonumber
\end{eqnarray}

Here, ACC, which is the percentage of correctly classified data in all the test data, indicates the model's performance.
$\Delta$ACC is the difference of the ACC values between the repaired model and the original model.
RR is the ratio of correctly classified data among the data misclassified by the original model before repair. 
BR is the ratio of misclassified data among the data correctly classified by the original model.

It should be noted the impact of BR values is larger than that of RR as generally $|I_{\textrm{pos}}|$ is much larger than $|I_{\textrm{neg}}|$. For example, breaking 1\% of positive inputs and repairing 1\% of negative inputs mean the impact of regressions is very dominant. We also look at the number of broken and repaired (patched) samples to investigate the trade-off. In addition, we expect suppressing regressions, i.e., achieving stably low RR values, is a unique and effective feature of \textsc{NeuRecover}.

\subsubsection{Experimental Environment}
\label{subsubsec:environment}
\textsc{NeuRecover} and Arachne were implemented in Python 3.6.9, and the DNN training models were implemented in TensorFlow 2.4.1.

\textsc{NeuRecover} has several hyper-parameters.
PSO used $\phi_1=\phi_2=4.1$, and the maximum number of iterations is 100.
PSO uses a population size of 200.
In addition, the weight rate of our fitness function $\alpha$ is 1.
However, in the experiment for specific misclassified data, the weight rate $\alpha$ is 5.

\subsection{Research Questions}
\label{subsec:research_questions}
We conducted the experiment to evaluate the effectiveness of \textsc{NeuRecover} by answering the following research questions.
\begin{description}
    \item[RQ1] Are the design elements of \textsc{NeuRecover} beneficial?
    \begin{description}
        \item[RQ1-1] How does use of the training history affect the repair performance? 
        \item[RQ1-2] How do the other variations in the repair method affect the repair performance?
    \end{description}
    \item[RQ2] Is \textsc{NeuRecover} effective in controlling regressions in repair tasks?
    \item[RQ3] Is \textsc{NeuRecover} effective in controlling regressions in fine-grained repair tasks for specific failure types?
\end{description}

We investigate effectiveness of each technical feature in \textsc{NeuRecover} compared with the baseline method Arachne in RQ1. RQ1-1 is about the core feature of \textsc{NeuRecover} to make use of the training history and RQ1-2 covers the other algorithm improvements. RQ2 and RQ3 evaluate the repair performance of \textsc{NeuRecover}, including all the features, with Arachne and retraining. We consider basic repair tasks and fine-grained repair tasks that do not or do focus on failure types, i.e., labels, respectively. We expect \textsc{NeuRecover} is more effective in fine-grained repair tasks where we have tight requirements on what to repair and thus hints from the localization phase help avoid manipulating unnecessarily large number of weight values.

\begin{table*}[t]
    \caption{RQ1-1. Comparison about use of the training history}
    \label{tb:rq1_1_results}
    \begin{center}
        \begin{tabular}{|l|r|r|r|r|r|r|r|r|r|r|}
            \hline
            Impact & \multicolumn{3}{c|}{All Impact} & \multicolumn{3}{c|}{Without Diff} & \multicolumn{3}{c|}{Without Improved Data} \\ \hline
            Model & $\Delta$ACC & RR & BR & $\Delta$ACC & RR & BR & $\Delta$ACC & RR & BR \\ \hline \hline
			GT+8CN & -1.067 & \textbf{13.418} & 1.679 & \textbf{0.067} & 12.981 & 0.448 & 0.063 & 11.856 & \textbf{0.420} \\ \hline
			GT+V16 & \textbf{0.374} & 18.114 & \textbf{1.714} & -1.083 & \textbf{23.526} & 3.975 & -0.148 & 21.693 & 2.725 \\ \hline
			GT+V19 & \textbf{-1.015} & 8.476 & \textbf{9.158} & -2.007 & 8.302 & 10.857 & -3.702 & \textbf{10.305} & 15.716 \\ \hline
			C10+8CN & -2.105 & 5.976 & 4.978 & \textbf{0.322} & \textbf{6.276} & \textbf{1.750} & -0.248 & 4.844 & 2.065 \\ \hline
			C10+V16 & -0.090 & 16.727 & 4.081 & \textbf{0.743} & 16.121 & \textbf{2.978} & -0.735 & \textbf{21.202} & 6.069 \\ \hline
			C10+V19 & -1.420 & 8.629 & 8.777 & \textbf{-1.033} & 7.845 & \textbf{7.530} & -2.548 & \textbf{11.152} & 12.573 \\ \hline
			FM+8CN & \textbf{-0.020} & 6.161 & \textbf{0.684} & -0.720 & \textbf{10.589} & 1.930 & -0.343 & 9.208 & 1.353 \\ \hline
			FM+V16 & \textbf{0.160} & 15.074 & \textbf{1.310} & -0.433 & \textbf{23.128} & 2.748 & -0.213 & 19.717 & 2.168 \\ \hline
			FM+V19 & \textbf{-0.083} & 7.450 & \textbf{1.536} & -1.435 & 12.407 & 4.100 & -2.140 & \textbf{14.124} & 5.277 \\ \hline
        \end{tabular}
    \end{center}
\end{table*}

\begin{table}[t]
    \caption{RQ1-2 (partial). Comparison of fitness functions}
    \label{tb:rq1_2_results}
    \begin{center}
        \begin{tabular}{|l|r|r|r|r|r|r|}
            \hline
            Fitness & \multicolumn{3}{c|}{\textsc{NeuRecover}} & \multicolumn{3}{c|}{Arachne} \\ \hline
            Model & $\Delta$ACC & RR & BR & $\Delta$ACC & RR & BR \\ \hline \hline
			GT+8CN & -1.067 & \textbf{13.418} & 1.679 & \textbf{0.032} & 8.779 & \textbf{0.326} \\ \hline
			GT+V16 & \textbf{0.374} & \textbf{18.114} & \textbf{1.714} & 0.329 & 17.549 & 1.726 \\ \hline
			GT+V19 & \textbf{-1.015} & 8.476 & \textbf{9.158} & -5.780 & \textbf{12.656} & 21.600 \\ \hline
			C10+8CN & \textbf{-2.105} & 5.976 & \textbf{4.978} & -10.278 & \textbf{17.947} & 19.928 \\ \hline
			C10+V16 & \textbf{-0.090} & 16.727 & \textbf{4.081} & -7.958 & \textbf{37.529} & 18.723 \\ \hline
			C10+V19 & \textbf{-1.420} & 8.629 & \textbf{8.777} & -11.735 & \textbf{19.592} & 34.662 \\ \hline
			FM+8CN & \textbf{-0.020} & 6.161 & \textbf{0.684} & -5.570 & \textbf{28.136} & 9.080 \\ \hline
			FM+V16 & \textbf{0.160} & 15.074 & \textbf{1.310} & -9.380 & \textbf{39.973} & 14.196 \\ \hline
			FM+V19 & \textbf{-0.083} & 7.450 & \textbf{1.536} & -11.925 & \textbf{30.590} & 20.115 \\ \hline
        \end{tabular}
    \end{center}
\end{table}

\begin{table*}[t]
    \caption{RQ2. Comparison between \textsc{NeuRecover} and Arachne}
    \label{tb:rq2_results1}
    \begin{center}
        \begin{tabular}{|l|l|r|r||r|r|r|r|r|r|}
            \cline{5-10}
            \multicolumn{4}{c|}{} & \multicolumn{3}{c|}{\textsc{NeuRecover}} & \multicolumn{3}{c|}{Arachne} \\ \hline
            Datasets & Model Arch. & Epochs & orig ACC & ACC & RR & BR & ACC & RR & BR \\ \hline \hline
			GTSRB & 8-layer CNN & 5 & 95.893 & 94.662 & \textbf{11.856} & 1.817 & \textbf{94.751} & 10.262 & \textbf{1.649}  \\ \hline
			GTSRB & 8-layer CNN & 10 & 96.247 & 95.180 & \textbf{13.418} & 1.679 & \textbf{95.907} & 9.728 & \textbf{0.732} \\ \hline
			GTSRB & VGG16 & 5 & 87.201 & \textbf{87.223} & \textbf{27.482} & 4.064 & 86.158 & 15.949 & \textbf{3.533} \\ \hline
			GTSRB & VGG16 & 10 & 89.733 & \textbf{90.107} & \textbf{18.114} & 1.714 & 89.515 & 8.482 & \textbf{1.205} \\ \hline
			GTSRB & VGG19 & 5 & 51.653 & \textbf{51.247} & 5.476 & \textbf{5.909} & 43.452 & \textbf{12.696} & 27.744 \\ \hline
			GTSRB & VGG19 & 10 & 53.830 & \textbf{52.815} & 8.476 & \textbf{9.158} & 45.283 & \textbf{13.062} & 27.083 \\ \hline
			CIFAR10 & 8-layer CNN & 5 & 67.635 & \textbf{67.643} & 0.200 & \textbf{0.088} & 61.045 & \textbf{11.414} & 15.018 \\ \hline
			CIFAR10 & 8-layer CNN & 10 & 74.580 & \textbf{72.475} & 5.976 & \textbf{4.978} & 67.495 & \textbf{13.025} & 13.925 \\ \hline
			CIFAR10 & VGG16 & 5 & 78.885 & \textbf{79.158} & 4.289 & \textbf{0.805} & 77.538 & \textbf{15.495} & 5.882 \\ \hline
			CIFAR10 & VGG16 & 10 & 80.880 & \textbf{80.790} & \textbf{16.727} & 4.081 & 79.918 & 10.915 & \textbf{3.803} \\ \hline
			CIFAR10 & VGG19 & 5 & 55.905 & \textbf{55.030} & 6.702 & \textbf{6.830} & 42.273 & \textbf{19.359} & 39.651 \\ \hline
			CIFAR10 & VGG19 & 10 & 57.738 & \textbf{56.318} & 8.629 & \textbf{8.777} & 43.475 & \textbf{18.411} & 38.182  \\ \hline
			Fashion-MNIST & 8-layer CNN & 5 & 89.495 & \textbf{89.500} & 0.720 & \textbf{0.078} & 85.623 & \textbf{14.730} & 6.059 \\ \hline
			Fashion-MNIST & 8-layer CNN & 10 & 90.555 & \textbf{90.535} & 6.161 & \textbf{0.684} & 87.028 & \textbf{16.494} & 5.620 \\ \hline
			Fashion-MNIST & VGG16 & 5 & 89.923 & \textbf{90.070} & 12.225 & \textbf{1.278} & 87.368 & \textbf{16.294} & 4.715 \\ \hline
			Fashion-MNIST & VGG16 & 10 & 91.110 & \textbf{91.270} & \textbf{15.074} & 1.310 & 91.140 & 5.746 & \textbf{0.524} \\ \hline
			Fashion-MNIST & VGG19 & 5 & 82.453 & \textbf{81.313} & 8.973 & \textbf{3.293} & 70.335 & \textbf{27.305} & 20.515 \\ \hline
			Fashion-MNIST & VGG19 & 10 & 83.850 & \textbf{83.768} & 7.450 & \textbf{1.536} & 70.885 & \textbf{26.252} & 20.526 \\ \hline
        \end{tabular}
    \end{center}
\end{table*}

\begin{table*}[t]
    \caption{RQ2. Comparison between \textsc{NeuRecover} and Retraining (Partial)}
    \label{tb:rq2_results2}
    \begin{center}
        \begin{tabular}{|l|l|r|r||r|r|r|r|r|r|}
            \cline{5-10}
            \multicolumn{4}{c|}{} & \multicolumn{3}{c|}{\textsc{NeuRecover}} & \multicolumn{3}{c|}{Retraining} \\ \hline
            Datasets & Model Arch. & Epochs & orig ACC & ACC & RR & BR & ACC & RR & BR \\ \hline \hline
			GTSRB & 8-layer CNN & 10 & 96.247 & 95.180 & 13.418 & 1.679 & \textbf{96.958} & \textbf{45.730} & \textbf{1.073} \\ \hline
			GTSRB & VGG16 & 10 & 89.733 & 90.107 & 18.114 & \textbf{1.714} & \textbf{91.059} & \textbf{29.905} & 1.993 \\ \hline
			GTSRB & VGG19 & 10 & 53.830 & 52.815 & 8.476 & 9.158 & \textbf{55.689} & \textbf{12.281} & \textbf{7.090} \\ \hline
			CIFAR10 & 8-layer CNN & 10 & 74.580 & 72.475 & 5.976 & \textbf{4.978} & \textbf{76.032} & \textbf{38.147} & 11.065 \\ \hline
			CIFAR10 & VGG16 & 10 & 80.880 & 80.790 & 16.727 & \textbf{4.081} & \textbf{83.715} & \textbf{36.172} & 5.061 \\ \hline
			CIFAR10 & VGG19 & 10 & 57.738 & 56.318 & 8.629 & 8.777 & \textbf{59.787} & \textbf{13.759} & \textbf{6.521} \\ \hline
			Fashion-MNIST & 8-layer CNN & 10 & 90.555 & 90.535 & 6.161 & \textbf{0.684} & \textbf{91.060} & \textbf{35.220} & 3.122 \\ \hline
			Fashion-MNIST & VGG16 & 10 & 91.110 & 91.270 & 15.074 & \textbf{1.310} & \textbf{92.045} & \textbf{30.777} & 1.978 \\ \hline
			Fashion-MNIST & VGG19 & 10 & 83.850 & 83.768 & 7.450 & \textbf{1.536} & \textbf{85.172} & \textbf{19.720} & 2.224 \\ \hline
        \end{tabular}
    \end{center}
\end{table*}

\begin{table*}[t]
    \caption{RQ3. Comparison of label-wise repair between \textsc{NeuRecover}, Arachne and Retraining}
    \label{tb:rq3_results}
    \begin{center}
        \begin{tabular}{|l|r|r|r|r|r|r|r|r|r|r|}
            \cline{3-11}
            \multicolumn{2}{c|}{} & \multicolumn{3}{c|}{\textsc{NeuRecover}} & \multicolumn{3}{c|}{Arachne} & \multicolumn{3}{c|}{Retraining} \\ \hline
            Model & LW-\#neg & $\Delta$ACC & LW-RR & BR & $\Delta$ACC & LW-RR & BR & $\Delta$ACC & LW-RR & BR \\ \hline \hline
            C10+8CN & 448.0 & \textbf{-0.405} & 12.444 & \textbf{1.759} & -2.270 & \textbf{37.612} & 6.480 & -0.480 & 35.938 & 12.672 \\ \hline
            C10+V16 & 339.3 & 0.333 & 4.643 & \textbf{0.489} & -0.995 & \textbf{33.161} & 3.725 & \textbf{1.227} & 17.318 & 4.944 \\ \hline
            C10+V19 & 603.8 & -0.155 & 5.880 & \textbf{2.035} & -7.317 & \textbf{49.358} & 23.708 & \textbf{0.082} & 9.400 & 8.547 \\ \hline
            FM+8CN & 292.8 & -0.165 & 8.198 & \textbf{0.546} & -1.383 & 19.129 & 2.888 & \textbf{0.025} & \textbf{22.545} & 3.131 \\ \hline
            FM+V16 & 243.3 & \textbf{0.135} & 1.953 & \textbf{0.272} & -0.235 & 9.455 & 0.732 & -0.313 & \textbf{15.313} & 2.814 \\ \hline
            FM+V19 & 414.8 & -0.002 & 1.989 & \textbf{0.200} & -9.188 & \textbf{70.283} & 15.729 & \textbf{0.200} & 9.222 & 2.289 \\ \hline
        \end{tabular}
    \end{center}
\end{table*}

\subsection{Results}
\label{subsec:results}

\subsubsection{{\bf RQ1-1. How does use of the training history affect the repair performance?}}
\label{subsubsec:rq2}

The key idea of \textsc{NeuRecover} is to make use of the training history in the localization phase. Specifically, we focused on weights whose values changed a lot and weights that affected improved data (Section \ref{subsec:our_localize}. In RQ 1-1, we investigate how these two points work.

The results are shown in Table \ref{tb:rq1_1_results}. The best values for each metric ($\Delta$ACC, RR, BR) are shown in bold. The rightmost column (\textit{Without Improved Data}) has large RR values but sometimes also large BR values, resulting in less ACC values. This point suggests that the idea to avoid manipulating weights that contributed improvement in the training history is working as expected. The left and center columns have comparative scores ($All Impact$ and $Without Diff$) but the left tends to have low BR. We claim that the proposed $All Impact$ is more stable when we are concerned about regressions.

\begin{itembox}[l]{Answer to RQ1-1}
The proposed ideas to use the training history in \textsc{NeuRecover} contribute to suppress regressions in DNN repair.
\end{itembox}

\subsubsection{{\bf RQ1-2. How do the other variations in the repair method affect the repair performance?}}
\label{subsubsec:rq3}
We had a few improvements in \textsc{NeuRecover} compared with the baseline Arachne implementation. Specifically, we included many layers as the target of repair, changed the way of sampling to calculate the fitness to reflect the loss, and changed the fitness function to be relative to the number of samples. As RQ1-2, we experimentally confirmed these changes some or less contribute to the repair performance. 

We omit the concrete results for the first two aspects due to space limitation. Table \ref{tb:rq1_2_results} shows the results on the third point, the fitness function, which had the largest impact on the repair performance. The modified fitness function contributes to suppress the regressions (low BR), resulting in better overall accuracy (high $\Delta$ACC).

\begin{itembox}[l]{Answer to RQ1-2}
The algorithm improvements in \textsc{NeuRecover}, especially in the fitness function, contribute to the DNN repair performance for suppressing regressions.
\end{itembox}

\subsubsection{{\bf RQ2. Is \textsc{NeuRecover} effective in controlling regressions in repair tasks?}}
\label{subsubsec:rq1}
\begin{figure}[t]
  \begin{center}
    \includegraphics[width=.7\linewidth]{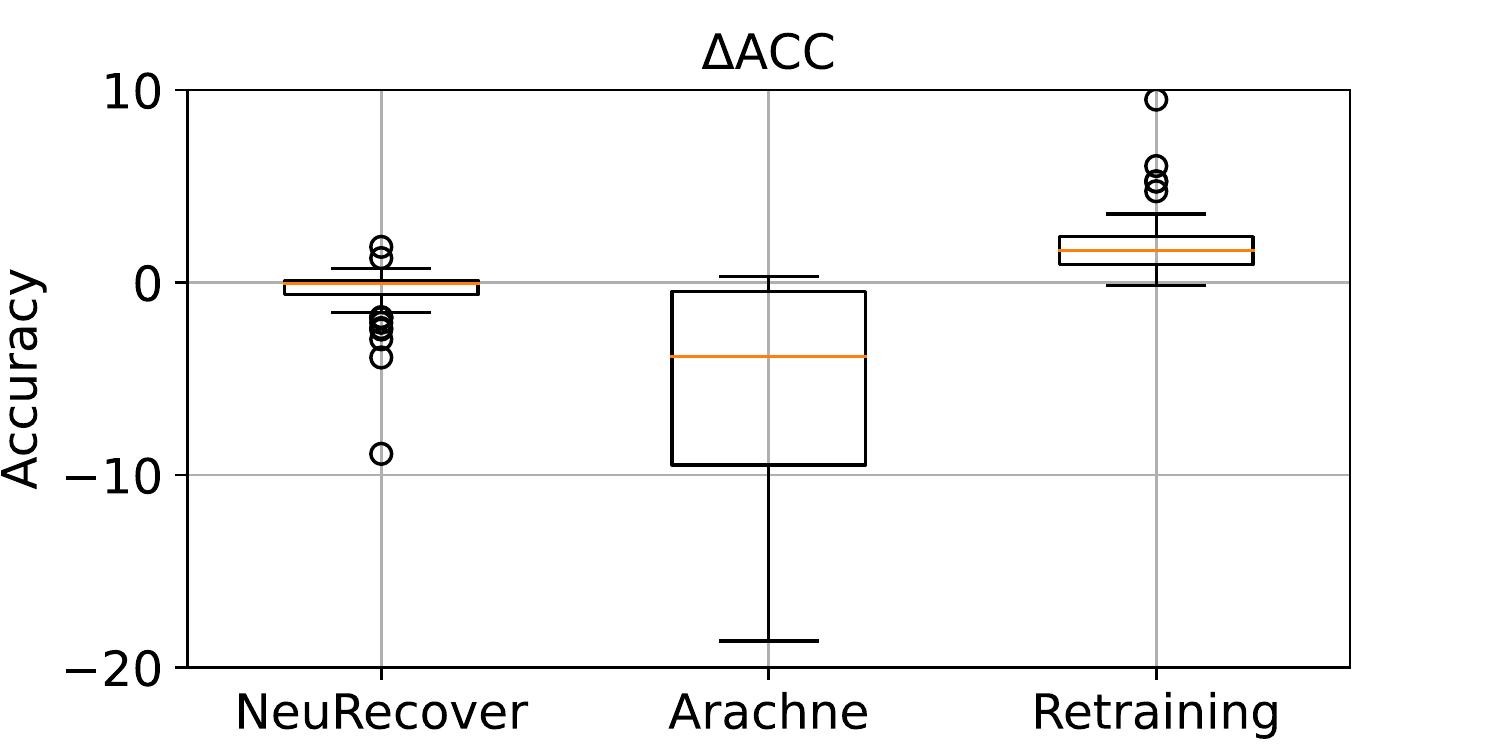}
    \includegraphics[width=.7\linewidth]{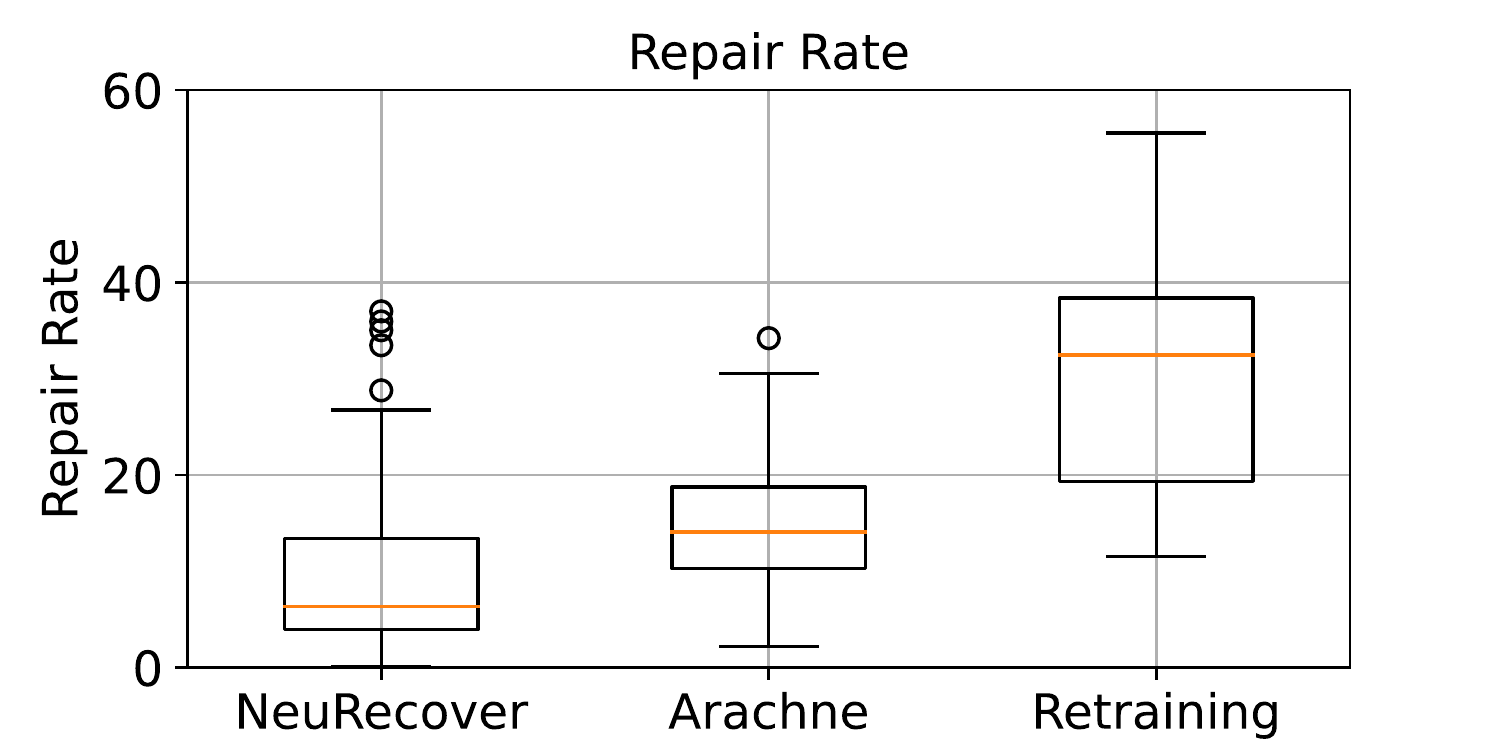}
    \includegraphics[width=.7\linewidth]{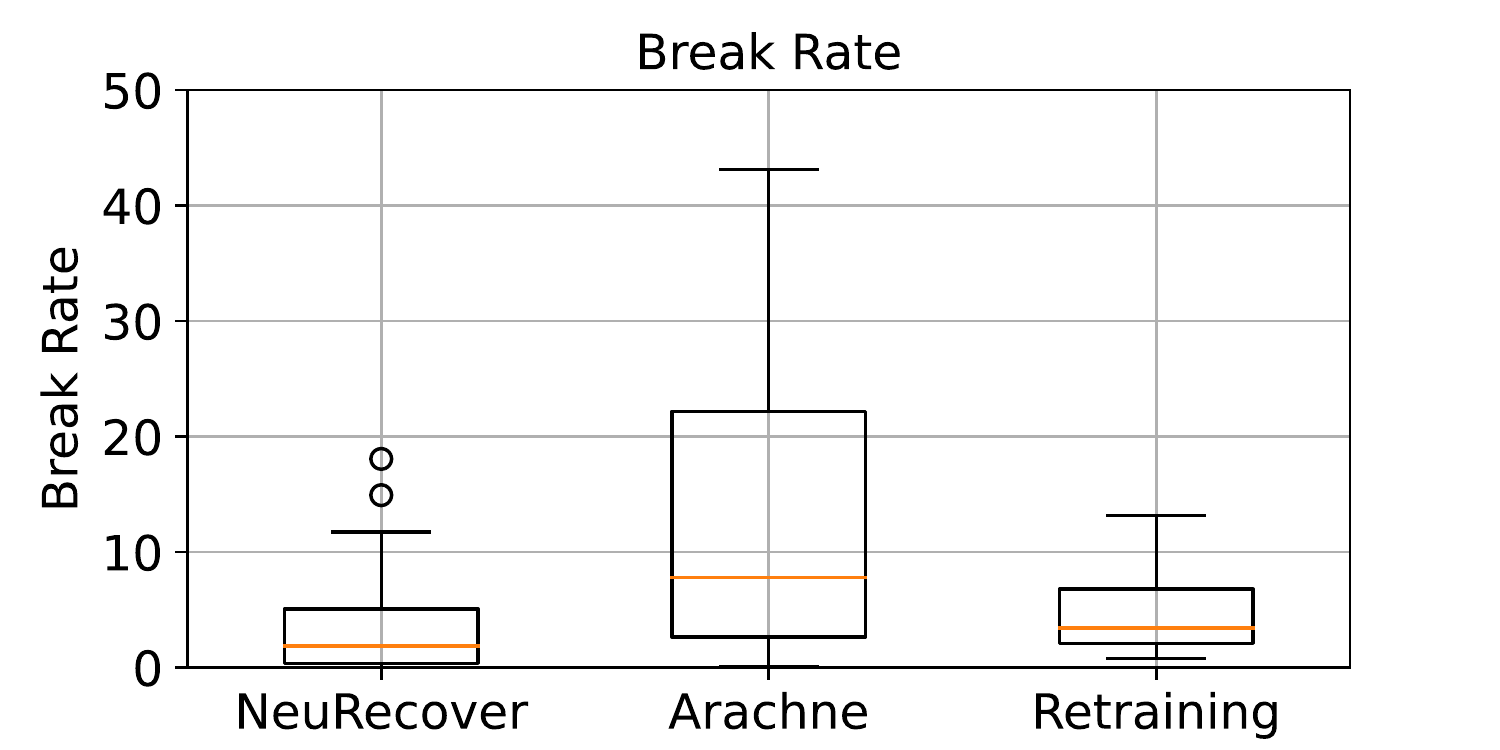}
  \end{center}
  \caption{Comparison of \textsc{NeuRecover}, Arachne and Retraining (the increase rate of ACC, RR and BR)}
  \label{fig:acc_rr_br_box_plot}
\end{figure}

We compared the repair performance of \textsc{NeuRecover}, i.e., ACC, RR, and BR, with the baseline Arachne and also with Retrain. The total results are summarized in Figure \ref{fig:acc_rr_br_box_plot} and the detailed comparisons with each target are shown in Tables \ref{tb:rq2_results1} and \ref{tb:rq2_results2}, respectively.

We start with discussion on \textsc{NeuRecover} and Arachne in in Table \ref{tb:rq2_results1}. \textsc{NeuRecover} achieved better (lower) BR scores than Arachne. BR scores of Arachne are unstable and often very high (many over 10\% and at worst even almost 40\%). BR scores of \textsc{NeuRecover} are stably low (below 10\%). \textsc{NeuRecover} often had less than a quarter, even a tenth in some cases, number of regressions compared with Arachne. As a result, ACC values are better in \textsc{NeuRecover} in most cases. Arachne achieved better (higher) RR scores in most cases but the regressions negated the improvement. These points are also summarized in the box plot of Figure \ref{fig:acc_rr_br_box_plot}.

Comparison with retraining is shown in Table \ref{tb:rq2_results2}. This table is partial only for Epochs=10 due to space limitation as the omitted parts had very similar tendency. In general, retraining shows better repair performance though \textsc{NeuRecover} keeps better (lower) BR values. One hypothesis is that the potential of search-based repair is not in repairing any failed inputs but in repairing specific failed inputs. Retraining can take the freedom to pick up ``easy-to-fix'' failed inputs and sufficiently works. This point is investigated in the following RQ3.

\begin{itembox}[l]{Answer to RQ2}
\textsc{NeuRecover} outperforms the baseline, Arachne, by stably suppressing regressions. Retraining is appropriate when the repair requirements are not tight, i.e., when improvements for any failed inputs are appreciated and some regressions are accepted. \textsc{NeuRecover} is a good option when the number of regressions is critical. 
\end{itembox}

\subsubsection{{\bf RQ3. Is \textsc{NeuRecover} effective in controlling regressions in fine-grained repair tasks for specific failure types?}}
\label{subsubsec:rq6}

We evaluated repair performance in terms of fine-grained control. Specifically, we consider popular scenarios in which a specific type of failures occur too frequently and we want to repair it. We picked up models with epochs=10 and defined the repair target by investigating the model performance. For CIFAR-10, the repair target was set as misclassification of label 3 to 5 (cat to dog). For Fashion-MNIST, the target was misclassification of label 6 to 0 (shirts to T-shirts). Both are representatives of confusing (visually close) labels. GTSRB was not included as it has many labels and the number of data for each specific failure type is too small.

Table \ref{tb:rq3_results} shows the results. The negative data (failed inputs) are considered for the specific label (in the label-wise way: LW). Thus, LW-\#neg and LW-RR denote the number of negative (failed) inputs and the repair rate for the label, respectively. \textsc{NeuRecover} outperforms Arachne with stably low BR (less than 2\%). Retraining shows worse ACC and BR compared with the case of RQ2 (Table \ref{tb:rq2_results2}). As the result, \textsc{NeuRecover} showed good controllability with stably low BR, with in many cases a tenth, at most a quarter, number of regressions compared with retraining.

\begin{figure*}
    \centering
    \begin{subfigmatrix}{2}
    \subfigure[CIFAR10/VGG16]{\includegraphics{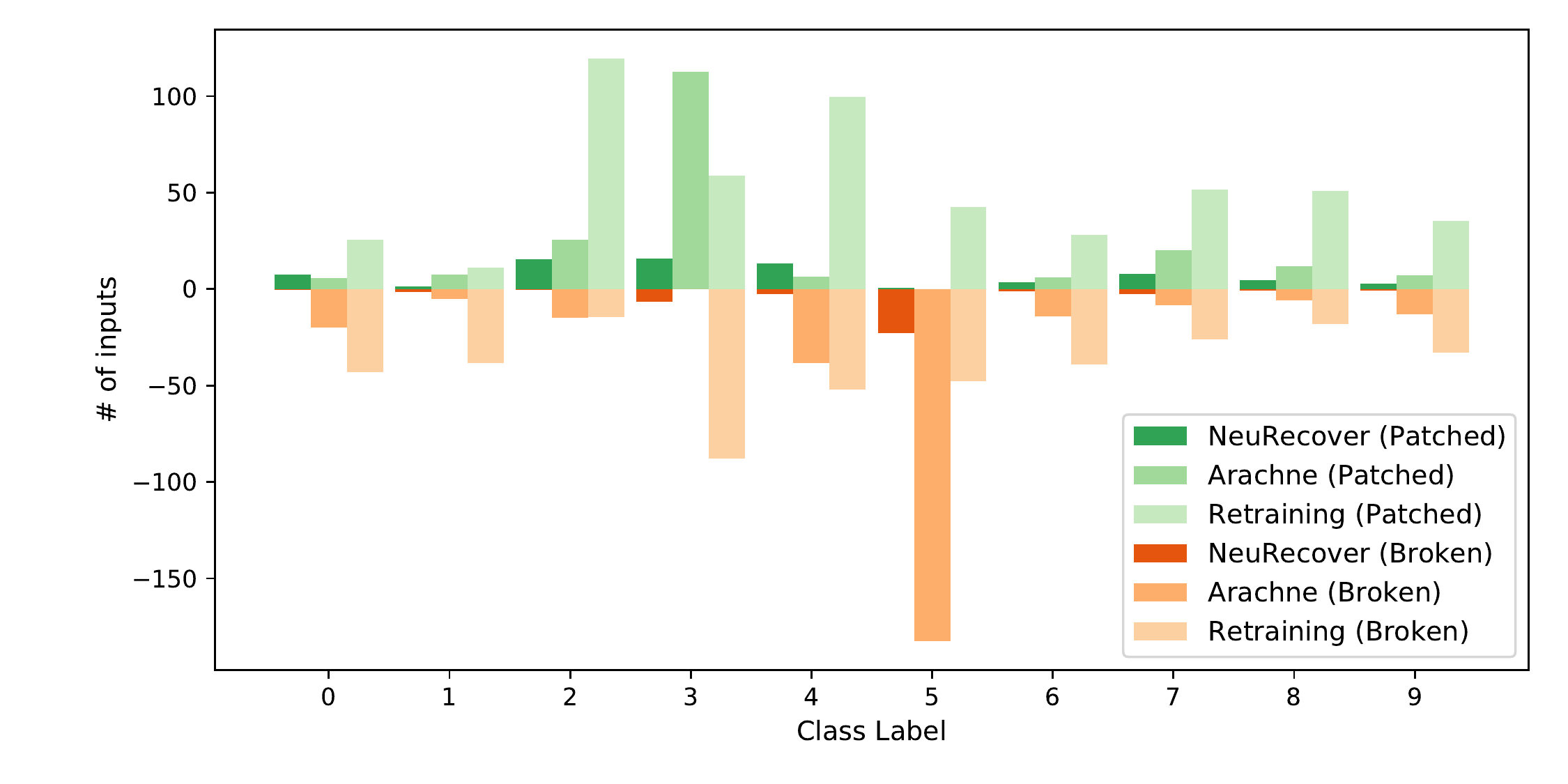}}
    \subfigure[Fashion-MNIST/VGG16]{\includegraphics{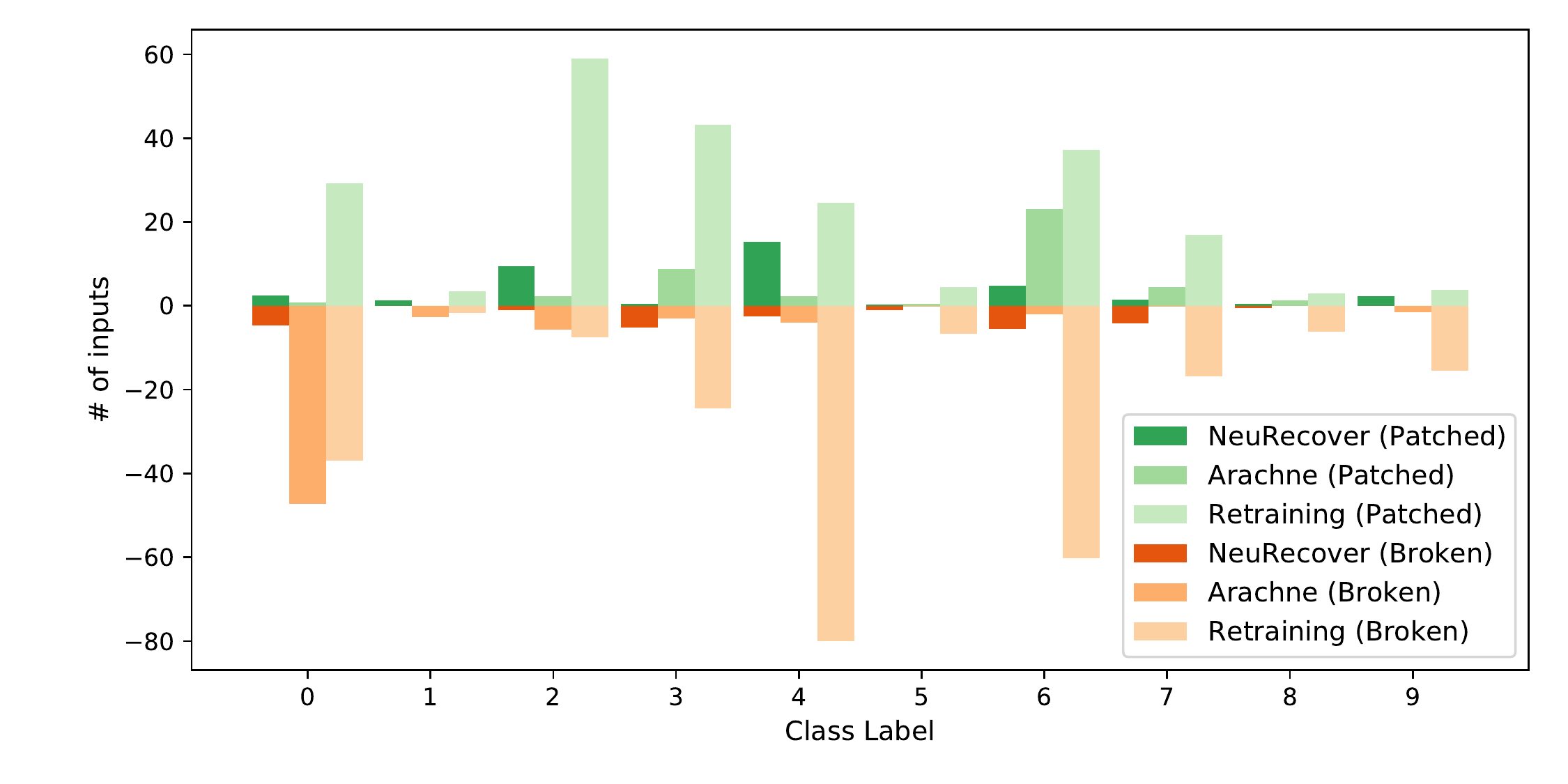}}
    \end{subfigmatrix}
    \caption{RQ3. Label-wise repair performance}
    \label{fig:label-wise-detail}
\end{figure*}

Figure \ref{fig:label-wise-detail} shows detailed label-wise repair performance. We picked up results for VGG16 with CIFAR-10 and Fashion-MNIST as the other results shared similar characteirstics. The figures show how the prediction results were changed - patched or broken. Although \textsc{NeuRecover} shows modest numbers of patched inputs, it keeps numbers of broken inputs low. Arachne tends to repair the target label a lot but instead has radical regressions in another label (label 5 for CIFAR-10 and label 0 for Fashion-MNIST). 

For retraining, regressions, or broken inputs, appear in various labels. It is notable that retraining has \textit{shuffling effect}: many improvements and regressions occur at the same time even for the same label, e.g., label 4 for Fashion-MNIST. This behavior is very critical when we consider intensive assurance activities to check risks of failed inputs even if the total accuracy remains similar or better.

\begin{itembox}[l]{Answer to RQ3}
\textsc{NeuRecover} outperforms the baseline of Arachne by stably suppressing regressions also in repairing specific failure types. \textsc{NeuRecover} outperforms retraining in suppressing regressions. Retraining tends to have more diverse regressions often with large shuffling of success and failure inputs. 
\end{itembox}

\subsection{Discussion}
\label{subsec:discussion}

All the experimental results suggest the key benefits of \textsc{NeuRecover} lie in its controllability in repair outcome, specifically, the capability to suppress regressions. On the other hand, repair performance in terms of RR is modest compared with Arachne or retraining. We can say \textsc{NeuRecover} is conservative not to make destructive changes that introduce large regressions even if they introduce more improvements.

We argue this feature is significant when the impact of failures is large in safety-critical or quality-sensitive applications. In such cases, engineers and stakeholders are more careful to check whether each of failed inputs is acceptable in terms of safety, ethics, or other risks that affect trust on the target system. Regressions, even with a larger number of improvements, require costly recheck on the newly introduced failures. In this case, the conservative approach of \textsc{NeuRecover} easily leads to modest but acceptable updates.

$\Delta$ACC values were sometimes around zero or negative for all the methods, especially in experiments for RQ3 with tight repair requirements. This fact suggests that the repair tasks intrinsically involve trade-offs and we cannot have a silver ballet to have improvements without any regressions. We thus assume that achieving high ACC values for whole the dataset alone is not the goal and it is necessary to argue impacts of specific success or failure types.

We had an informal workshop with industry practitioners from more than ten companies to discuss significance of considering fine-grained repair tasks such as the label-wise one for RQ3. All the practitioners agreed with the significance and showed concrete examples of repair requirements as follows.
\begin{itemize}
\item Specific failure types with worst performance should be fixed (the experimental setting of RQ3).
\item Some labels are more significant than others, e.g., misrecognizing ``stop'' signs to something else is very critical.
\item Some failure types are more critical than others, e.g., misrecognizing something to ''go ahead'' signs is very critical.
\end{itemize}
The conservative approach of \textsc{NeuRecover} has potentials to deal with such fine-grained requirements as partially shown in experiments for RQ3.

\begin{itembox}[l]{Applicability}
\textsc{NeuRecover} is suitable when failures are critical and engineers have costly tasks to check regressions for safety or trust assurance and/or when there are fine-grained requirements to prioritize labels or failure types.
\end{itembox}

\subsection{Threats to Validity}
\label{subsec:threats}
The core threat to the internal validity is that we did not include experiments over different optimization methods.
In addition to PSO, we can consider using many other optimization methods, such as genetic algorithms and gradient descent.

The core threat to the external validity is the quantity of evaluation objects in the experiments.
The experiment uses three datasets and three CNN architectures for image classification.
It will be necessary to increase the number and types of datasets, models, and ML tasks to show that our technique is not dependent on a specific experimental object.

\section{Related Work}
\label{sec:relatedworks}

Many works on DNN testing and debugging have been inspired by software engineering techniques.
One of these techniques is automated program repair (APR), which generates patches that make buggy programs pass all test cases.
Some APR techniques achieve high repair performance by using the code editing history as a hint for repair \cite{long2016automatic,saha2017elixir}.
Spectrum-based fault localization, which is a part of APR, provides a score for suspiciousness by regarding program elements executed more frequently in failed test cases as more suspicious.
In the same way for fault localization, the code editing history can be used to improve the accuracy of localizing faults \cite{sohn2017fluccs,li2019deepfl}.
Our approach is motivated by the success of many history-based debugging methods.

Retraining is the most popular approach to fixing DNNs.
Studies of test data generation techniques for DNNs have shown that retraining with adversarial examples generated as test data can improve robustness \cite{sensei2020gao}.
Several techniques have been proposed to fix specified failures in general, not just adversarial examples.
\textit{Few-Shot Guided Mix} (FSGMix) \cite{ren2020few} is an augmentation-based repair technique that augments retraining data with the guidance of limited failure data.
Srivastava et al. proposed a model learning scheme that adds a compatibility penalty to the loss function \cite{srivastava2020empirical}.
Yan et al. also proposed a retraining method that suppresses negative flip by adding penalty term based on model distillation to the loss function\cite{yan2021positive}.
MODE is a debugging method for DNNs that works by identifying the features that are most affected by failed tests and generating inputs that are focused on those features by using a generative adversarial network (GAN).

We referred to Arachne \cite{arachne2019Sohn} as the baseline method with the same approach of directly manipulating DNN weight parameters. Apricot \cite{zhang2019apricot} is another technique to obtain hints from different versions of DNNs created by using subsets of the training data. This approach rather captures how to fix the behavior for negative input data.

Continual Learning and similar techniques can be referred to as an effort to maintain deep learning model performance \cite{HASSABIS2017245}, \cite{doi:10.2200/S00832ED1V01Y201802AIM037}. 
They aim at reducing catastrophic forgetting (interference) \cite{FRENCH1999128}, \cite{MCCLOSKEY1989109} enabling a deep learning model to learn a new task keeping previously learned tasks \cite{THRUN199525}. 
De Lange et al. classified Continual Learning into three categories \cite{HASSABIS2017245}: \textit{Replay methods}, \textit{Regularization-based methods} and \textit{Parameter-isolation methods}.
Among them, several works are in common with \textsc{Neurecover} in an aspect of modifying specific parameters with intention to maintain performance.

\textit{Parameter-isolation method} include techniques that incorporate isolating parameters such as branching multiple versions of a DNN model corresponding to each task or fixing specific weights during training \cite{9349197}. Mallya and Lazebnik proposed PackNet to fix important parameters for one task, updating only the rest of the parameters. The task-oriented updates are repeated on sequential tasks \cite{Mallya_2018_CVPR}.

Each \textit{Parameter-isolation method} above is similar to \textsc{Neurecover} because \textsc{Neurecover} extracts parameters concerned to classification faults (fault localization steps); however, \textsc{Neurecover} prevents a deep learning model from degrading performance with a single task. On the other hand, works related to Continual Learning attempt to maintain the model performance in regard of different tasks.

\section{Conclusion}
\label{sec:summary}
In this paper, we have presented a novel DNN repair method \textsc{NeuRecover} by using the training history. The proposed method outperforms the existing repair method with the same approach of search-based repair owing to the capability to stably suppress regressions. We also demonstrated our method is especially effective when the repair requirements are tight by requesting to fix specific failure types and to avoid regressions. The presented approach is suitable for safety-critical or quality-sensitive applications that require intensive assurance activities including risk evaluation of failure cases as well as consideration of fine-grained requirements to prioritize labels or failure types. We believe this work demonstrated the significant first-step for fine-grained, regression-aware, and controllable engineering of DNNs.

The following issues are future works for \textsc{NeuRecover}.
\begin{itemize}
    \item Implementation of a technique for identifying suspicious weights of convolutional layers
    \item Study on repair techniques for other DNN models besides the image classification problem
\end{itemize}

\hl{
\section*{Acknowledgment}
    This work was partly supported by JST-Mirai Program Grant Number JPMJMI20B8, Japan.
}

\bibliographystyle{IEEEtran}
\bibliography{references}

\begin{thebibliography}{10}
\providecommand{\url}[1]{#1}
\csname url@samestyle\endcsname
\providecommand{\newblock}{\relax}
\providecommand{\bibinfo}[2]{#2}
\providecommand{\BIBentrySTDinterwordspacing}{\spaceskip=0pt\relax}
\providecommand{\BIBentryALTinterwordstretchfactor}{4}
\providecommand{\BIBentryALTinterwordspacing}{\spaceskip=\fontdimen2\font plus
\BIBentryALTinterwordstretchfactor\fontdimen3\font minus
  \fontdimen4\font\relax}
\providecommand{\BIBforeignlanguage}[2]{{%
\expandafter\ifx\csname l@#1\endcsname\relax
\typeout{** WARNING: IEEEtran.bst: No hyphenation pattern has been}%
\typeout{** loaded for the language `#1'. Using the pattern for}%
\typeout{** the default language instead.}%
\else
\language=\csname l@#1\endcsname
\fi
#2}}
\providecommand{\BIBdecl}{\relax}
\BIBdecl

\bibitem{hinton2012deep}
G.~Hinton, L.~Deng, D.~Yu, G.~E. Dahl, A.-r. Mohamed, N.~Jaitly, A.~Senior,
  V.~Vanhoucke, P.~Nguyen, T.~N. Sainath \emph{et~al.}, ``Deep neural networks
  for acoustic modeling in speech recognition: The shared views of four
  research groups,'' \emph{IEEE Signal processing magazine}, vol.~29, no.~6,
  pp. 82--97, 2012.

\bibitem{cho2014learning}
K.~Cho, B.~van Merri{\"e}nboer, C.~Gulcehre, D.~Bahdanau, F.~Bougares,
  H.~Schwenk, and Y.~Bengio, ``Learning phrase representations using rnn
  encoder--decoder for statistical machine translation,'' in \emph{Proceedings
  of the 2014 Conference on Empirical Methods in Natural Language Processing
  (EMNLP)}, 2014, pp. 1724--1734.

\bibitem{ren2015faster}
S.~Ren, K.~He, R.~Girshick, and J.~Sun, ``Faster r-cnn: towards real-time
  object detection with region proposal networks,'' in \emph{Proceedings of the
  28th International Conference on Neural Information Processing Systems-Volume
  1}, 2015, pp. 91--99.

\bibitem{tang2015document}
D.~Tang, B.~Qin, and T.~Liu, ``Document modeling with gated recurrent neural
  network for sentiment classification,'' in \emph{Proceedings of the 2015
  conference on empirical methods in natural language processing}, 2015, pp.
  1422--1432.

\bibitem{schroff2015facenet}
F.~Schroff, D.~Kalenichenko, and J.~Philbin, ``Facenet: A unified embedding for
  face recognition and clustering,'' in \emph{Proceedings of the IEEE
  conference on computer vision and pattern recognition}, 2015, pp. 815--823.

\bibitem{litjens2017survey}
G.~Litjens, T.~Kooi, B.~E. Bejnordi, A.~A.~A. Setio, F.~Ciompi, M.~Ghafoorian,
  J.~A. Van Der~Laak, B.~Van~Ginneken, and C.~I. S{\'a}nchez, ``A survey on
  deep learning in medical image analysis,'' \emph{Medical image analysis},
  vol.~42, pp. 60--88, 2017.

\bibitem{chen2015deepdriving}
C.~Chen, A.~Seff, A.~Kornhauser, and J.~Xiao, ``Deepdriving: Learning
  affordance for direct perception in autonomous driving,'' in
  \emph{Proceedings of the IEEE international conference on computer vision},
  2015, pp. 2722--2730.

\bibitem{julian2016policy}
K.~D. Julian, J.~Lopez, J.~S. Brush, M.~P. Owen, and M.~J. Kochenderfer,
  ``Policy compression for aircraft collision avoidance systems,'' in
  \emph{2016 IEEE/AIAA 35th Digital Avionics Systems Conference (DASC)}.\hskip
  1em plus 0.5em minus 0.4em\relax IEEE, 2016, pp. 1--10.

\bibitem{amershi2019software}
S.~Amershi, A.~Begel, C.~Bird, R.~DeLine, H.~Gall, E.~Kamar, N.~Nagappan,
  B.~Nushi, and T.~Zimmermann, ``Software engineering for machine learning: A
  case study,'' in \emph{2019 IEEE/ACM 41st International Conference on
  Software Engineering: Software Engineering in Practice (ICSE-SEIP)}.\hskip
  1em plus 0.5em minus 0.4em\relax IEEE, 2019, pp. 291--300.

\bibitem{ishikawa2019engineers}
F.~Ishikawa and N.~Yoshioka, ``How do engineers perceive difficulties in
  engineering of machine-learning systems?-questionnaire survey,'' in
  \emph{2019 IEEE/ACM Joint 7th International Workshop on Conducting Empirical
  Studies in Industry (CESI) and 6th International Workshop on Software
  Engineering Research and Industrial Practice (SER\&IP)}.\hskip 1em plus 0.5em
  minus 0.4em\relax IEEE, 2019, pp. 2--9.

\bibitem{hamada2020guidelines}
K.~Hamada, F.~Ishikawa, S.~Masuda, M.~Matsuya, and Y.~Ujita, ``Guidelines for
  quality assurance of machine learning-based artificial intelligence,'' in
  \emph{SEKE2020: the 32nd International Conference on Software Engineering \&
  Knowledge Engineering}, 2020, pp. 335--341.

\bibitem{karpathy2017software}
\BIBentryALTinterwordspacing
A.~Karpathy, ``Software 2.0,'' 2017. [Online]. Available:
  \url{https://karpathy.medium.com/software-2-0-a64152b37c35}
\BIBentrySTDinterwordspacing

\bibitem{sculley2015hidden}
D.~Sculley, G.~Holt, D.~Golovin, E.~Davydov, T.~Phillips, D.~Ebner,
  V.~Chaudhary, M.~Young, J.-F. Crespo, and D.~Dennison, ``Hidden technical
  debt in machine learning systems,'' \emph{Advances in neural information
  processing systems}, vol.~28, pp. 2503--2511, 2015.

\bibitem{le2012systematic}
C.~Le~Goues, M.~Dewey-Vogt, S.~Forrest, and W.~Weimer, ``A systematic study of
  automated program repair: Fixing 55 out of 105 bugs for \$8 each,'' in
  \emph{2012 34th International Conference on Software Engineering
  (ICSE)}.\hskip 1em plus 0.5em minus 0.4em\relax IEEE, 2012, pp. 3--13.

\bibitem{saha2017elixir}
R.~K. Saha, Y.~Lyu, H.~Yoshida, and M.~R. Prasad, ``Elixir: Effective
  object-oriented program repair,'' in \emph{2017 32nd IEEE/ACM International
  Conference on Automated Software Engineering (ASE)}.\hskip 1em plus 0.5em
  minus 0.4em\relax IEEE, 2017, pp. 648--659.

\bibitem{noda2020experience}
K.~Noda, Y.~Nemoto, K.~Hotta, H.~Tanida, and S.~Kikuchi, ``Experience report:
  How effective is automated program repair for industrial software?'' in
  \emph{2020 IEEE 27th International Conference on Software Analysis, Evolution
  and Reengineering (SANER)}.\hskip 1em plus 0.5em minus 0.4em\relax IEEE,
  2020, pp. 612--616.

\bibitem{arachne2019Sohn}
\BIBentryALTinterwordspacing
J.~Sohn, S.~Kang, and S.~Yoo, ``Search based repair of deep neural networks,''
  \emph{arXiv preprint arXiv:1912.12463}, 2019. [Online]. Available:
  \url{http://arxiv.org/abs/1912.12463}
\BIBentrySTDinterwordspacing

\bibitem{pso1995James}
J.~Kennedy and R.~Eberhart, ``Particle swarm optimization,'' in \emph{Proc. of
  ICNN'95}, vol.~4.\hskip 1em plus 0.5em minus 0.4em\relax IEEE, 1995, pp.
  1942--1948.

\bibitem{pso2007Andreas}
\BIBentryALTinterwordspacing
A.~Windisch, S.~Wappler, and J.~Wegener, ``Applying particle swarm optimization
  to software testing,'' in \emph{Proc. of GECCO'07}.\hskip 1em plus 0.5em
  minus 0.4em\relax Association for Computing Machinery, 2007, pp. 1121--1128.
  [Online]. Available: \url{https://doi.org/10.1145/1276958.1277178}
\BIBentrySTDinterwordspacing

\bibitem{rumelhart1986learning}
D.~E. Rumelhart, G.~E. Hinton, and R.~J. Williams, ``Learning representations
  by back-propagating errors,'' \emph{nature}, vol. 323, no. 6088, pp.
  533--536, 1986.

\bibitem{gtsrb}
J.~Stallkamp, M.~Schlipsing, J.~Salmen, and C.~Igel, ``Man vs. computer:
  Benchmarking machine learning algorithms for traffic sign recognition,''
  \emph{Neural networks}, vol.~32, pp. 323--332, 2012.

\bibitem{cifar10}
A.~Krizhevsky, ``Learning multiple layers of features from tiny images,'' Tech.
  Rep., 2009.

\bibitem{fmnist}
H.~Xiao, K.~Rasul, and R.~Vollgraf. (2017) Fashion-mnist: a novel image dataset
  for benchmarking machine learning algorithms.

\bibitem{long2016automatic}
F.~Long and M.~Rinard, ``Automatic patch generation by learning correct code,''
  in \emph{Proceedings of the 43rd Annual ACM SIGPLAN-SIGACT Symposium on
  Principles of Programming Languages}, 2016, pp. 298--312.

\bibitem{sohn2017fluccs}
J.~Sohn and S.~Yoo, ``Fluccs: Using code and change metrics to improve fault
  localization,'' in \emph{Proceedings of the 26th ACM SIGSOFT International
  Symposium on Software Testing and Analysis}, 2017, pp. 273--283.

\bibitem{li2019deepfl}
X.~Li, W.~Li, Y.~Zhang, and L.~Zhang, ``Deepfl: Integrating multiple fault
  diagnosis dimensions for deep fault localization,'' in \emph{Proceedings of
  the 28th ACM SIGSOFT International Symposium on Software Testing and
  Analysis}, 2019, pp. 169--180.

\bibitem{sensei2020gao}
X.~Gao, R.~K. Saha, M.~R. Prasad, and A.~Roychoudhury, ``Fuzz testing based
  data augmentation to improve robustness of deep neural networks,'' in
  \emph{Proc. of ICSE'20}.\hskip 1em plus 0.5em minus 0.4em\relax IEEE, 2020,
  pp. 1147--1158.

\bibitem{ren2020few}
X.~Ren, B.~Yu, H.~Qi, F.~Juefei-Xu, Z.~Li, W.~Xue, L.~Ma, and J.~Zhao,
  ``Few-shot guided mix for dnn repairing,'' in \emph{Proc. of ICSME'20}.\hskip
  1em plus 0.5em minus 0.4em\relax IEEE, 2020, pp. 717--721.

\bibitem{srivastava2020empirical}
M.~Srivastava, B.~Nushi, E.~Kamar, S.~Shah, and E.~Horvitz, ``An empirical
  analysis of backward compatibility in machine learning systems,'' in
  \emph{Proc. of KDD'20}, 2020, pp. 3272--3280.

\bibitem{yan2021positive}
S.~Yan, Y.~Xiong, K.~Kundu, S.~Yang, S.~Deng, M.~Wang, W.~Xia, and S.~Soatto,
  ``Positive-congruent training: Towards regression-free model updates,'' in
  \emph{Proceedings of the IEEE/CVF Conference on Computer Vision and Pattern
  Recognition}, 2021, pp. 14\,299--14\,308.

\bibitem{zhang2019apricot}
H.~Zhang and W.~Chan, ``Apricot: A weight-adaptation approach to fixing deep
  learning models,'' in \emph{Proc. of ASE'19}.\hskip 1em plus 0.5em minus
  0.4em\relax IEEE, 2019, pp. 376--387.

\bibitem{HASSABIS2017245}
\BIBentryALTinterwordspacing
D.~Hassabis, D.~Kumaran, C.~Summerfield, and M.~Botvinick,
  ``Neuroscience-inspired artificial intelligence,'' \emph{Neuron}, vol.~95,
  no.~2, pp. 245--258, 2017. [Online]. Available:
  \url{https://www.sciencedirect.com/science/article/pii/S0896627317305093}
\BIBentrySTDinterwordspacing

\bibitem{doi:10.2200/S00832ED1V01Y201802AIM037}
\BIBentryALTinterwordspacing
Z.~Chen and B.~Liu, ``Lifelong machine learning, second edition,''
  \emph{Synthesis Lectures on Artificial Intelligence and Machine Learning},
  vol.~12, no.~3, pp. 1--207, 2018. [Online]. Available:
  \url{https://doi.org/10.2200/S00832ED1V01Y201802AIM037}
\BIBentrySTDinterwordspacing

\bibitem{FRENCH1999128}
\BIBentryALTinterwordspacing
R.~M. French, ``Catastrophic forgetting in connectionist networks,''
  \emph{Trends in Cognitive Sciences}, vol.~3, no.~4, pp. 128--135, 1999.
  [Online]. Available:
  \url{https://www.sciencedirect.com/science/article/pii/S1364661399012942}
\BIBentrySTDinterwordspacing

\bibitem{MCCLOSKEY1989109}
\BIBentryALTinterwordspacing
M.~McCloskey and N.~J. Cohen, ``Catastrophic interference in connectionist
  networks: The sequential learning problem,'' ser. Psychology of Learning and
  Motivation, G.~H. Bower, Ed.\hskip 1em plus 0.5em minus 0.4em\relax Academic
  Press, 1989, vol.~24, pp. 109--165. [Online]. Available:
  \url{https://www.sciencedirect.com/science/article/pii/S0079742108605368}
\BIBentrySTDinterwordspacing

\bibitem{THRUN199525}
\BIBentryALTinterwordspacing
S.~Thrun and T.~M. Mitchell, ``Lifelong robot learning,'' \emph{Robotics and
  Autonomous Systems}, vol.~15, no.~1, pp. 25--46, 1995, the Biology and
  Technology of Intelligent Autonomous Agents. [Online]. Available:
  \url{https://www.sciencedirect.com/science/article/pii/092188909500004Y}
\BIBentrySTDinterwordspacing

\bibitem{9349197}
M.~Delange, R.~Aljundi, M.~Masana, S.~Parisot, X.~Jia, A.~Leonardis,
  G.~Slabaugh, and T.~Tuytelaars, ``A continual learning survey: Defying
  forgetting in classification tasks,'' \emph{IEEE Transactions on Pattern
  Analysis and Machine Intelligence}, pp. 1--1, 2021.

\bibitem{Mallya_2018_CVPR}
A.~Mallya and S.~Lazebnik, ``Packnet: Adding multiple tasks to a single network
  by iterative pruning,'' in \emph{Proceedings of the IEEE Conference on
  Computer Vision and Pattern Recognition (CVPR)}, June 2018.

\end{thebibliography}
\end{document}